\documentclass{svproc}
\usepackage[utf8]{inputenc}
\usepackage[english]{babel}
\usepackage{graphicx}
\usepackage{xcolor}
\usepackage{subcaption}
\usepackage{ulem}
\usepackage{mwe}
\usepackage{array}
\usepackage{float}
\floatstyle{plaintop}
\restylefloat{table}
\usepackage{placeins}
\usepackage[labelfont=bf]{caption}
\usepackage{csquotes}

\usepackage{biblatex}
\usepackage{url}

\begin{document}
\mainmatter              
\title{Deep Learning for Cardiologist-level Myocardial Infarction Detection in Electrocardiograms}
\titlerunning{DL for MI Detection in ECGs}  
%


\author{AG\inst{1}\inst{2} \and EAH \inst{1}\inst{3}}
\author{Arjun Gupta\inst{1}\inst{2} \and Eliu Huerta\inst{1}\inst{3}\inst{4} \and Zhizhen Zhao\inst{1}\inst{2} \and Issam Moussa\inst{1}\inst{5}\inst{6}}
\authorrunning{Gupta et al.} 
\institute{NCSA Center for Artificial Intelligence Innovation, University of Illinois at Urbana-Champaign, Urbana, Illinois 61801, USA \and 
           Department of Electrical and Computer Engineering, University of Illinois at Urbana-Champaign, Urbana, Illinois 61801, USA \and 
           Department of Astronomy, University of Illinois at Urbana-Champaign, Urbana, Illinois 61801, USA \and
           Department of Physics, University of Illinois at Urbana-Champaign, Urbana, Illinois 61801, USA \and
           Beckman Institute for Advanced Science and Technology, University of Illinois at Urbana-Champaign, Urbana, Illinois 61801, USA \and
           Carle Illinois College of Medicine, University of Illinois at Urbana-Champaign, Illinois 61820, USA}

\maketitle              

\begin{abstract}
Myocardial infarction is the leading cause of death worldwide. In this paper, we design domain-inspired neural network models to detect myocardial infarction. First, we study the contribution of various leads. This systematic analysis, first of its kind in the literature, indicates that out of 15 ECG leads, data from the v6, vz, and ii leads are critical to correctly identify myocardial infarction. Second, we use this finding and adapt the ConvNetQuake neural network model—originally designed to identify earthquakes—to attain state-of-the-art classification results for myocardial infarction, achieving 99.43\% classification accuracy on a record-wise split, and 97.83\% classification accuracy on a patient-wise split. These two results represent cardiologist-level performance level for myocardial infarction detection after feeding only 10 seconds of raw ECG data into our model. Third, we show that our multi-ECG-channel neural network achieves cardiologist-level performance without the need of any kind of manual feature extraction or data pre-processing.
\end{abstract}
\section{Introduction}
Heart disease is the leading cause of death worldwide. Amongst patients with cardiovascular diseases, myocardial infarction is the main cause of death. In order to provide adequate healthcare support to patients who may experience this clinical event, it is essential to gather supportive evidence in a timely manner to help secure a correct diagnosis. 

Deep learning is the engine of the artificial intelligence renaissance we have experienced since the early 2010s~\cite{LeCun:Nature,SCHMIDHUBER201585}. Deep learning applications are ubiquitous in modern technologies, and there is a sense of urgency to better understand how these algorithms may be designed, trained and utilized in an optimal manner---an emergent area of research dubbed ``scientific machine learning"~\cite{doedl}.

The confluence of mathematical and statistical models, along with numerical simulations to inform the design and training of deep learning algorithms is a booming enterprise~\cite{maziar:DBLP,RAISSI2019686,Shen:2019vep,2019NatRP6}. Furthermore, the development of novel schemes to accelerate the training of deep learning models using hundreds to thousands of graphics processing units (GPUs) in high performance computing platforms has enabled the construction of more sophisticated neural network models, trained with TB-size data sets, that become more resilient to noise anomalies, and which exhibit state-of-the-art results for classification and regression analyses~\cite{eliu_nsf_dl,KHAN2019248,AIHPC,dglitcha:2017}. An additional advantage of these algorithms is that once the compute-intensive training stage is complete, they require minimal computational resources for inference analyses~\cite{Rebei:2018R,berg2018unified,Shen:ICASP2019,NIPS2016_6117,WEI2020135081,IsmailFawaz2019,geodf:2017a,geodf:2017b,AlvinC:2018,Dom:2018MNRAS,Dom:2018D,Dom:2018MNRAS,icecubenets,weistarclusters}. 
  
The combination of data fusion and machine learning exhibit great promise to enable innovation in healthcare. However, in order to realize this expectation, it is essential to first ascertain what AI architectures and data sets provide the most critical information to address a given challenge. In the case of diagnosis of myocardial injury, we require a number of supportive evidence in terms of typical symptoms, electrocardiographic (ECG) changes, imaging evidence that indicates new loss of viable myocardium or new regional wall motion abnormality, etc. Along this line of thought, in this article we:

\begin{enumerate}
    \item Fill a void in the literature regarding the use of ECG time-series data to identify heart conditions, i.e., in our literature review, we were unable to find a convincing line of argumentation regarding the choice of ECG data used in previous machine learning studies to identify heart conditions. Given that the choices in previous analyses are arbitrary, we present a detailed analysis of the contribution of each ECG lead for the identification of heart conditions.
    \item Illustrate the benefits of transferring knowledge between disparate areas of research that are threaded by a common theme, namely, the use of multiple data channels of information to enhance our confidence in the prediction of deep learning models. In this particular case, we adapt a neural network model that was designed to identify earthquakes by feeding into the model multiple channels of information. In our case, we first identify the ECG leads that provide the highest confidence to tell apart healthy from unhealthy hearts, and then re-design the  neural network model to process simultaneously the top three ECG leads that provide the highest classification results.
    \item Show that feature engineering is not necessary to curate the ECG data used to train, validate and test our neural network models, which currently represent cardiologist state-of-the-art performance for myocardial infarction detection. 
\end{enumerate}

\section{Related Work}

Myocardial infarction detection on the PTB database is a problem that has been studied extensively in the past. \cite{acharya} were one of the first to develop a convolutional neural network for the automatic detection of myocardial infarction. This was done on both noisy and de-noised electrocardiogram signals, without any manual feature extraction. \cite{safdarian} used two features-the T-wave integral and the total integral-for the localization and detection of myocardial infarction. Using these two features and various statistical learning techniques such as artificial neural networks, probabilistic neural networks, k-nearest neighbors, multi-layer perceptrons, and naive-Bayes classifiers, \cite{safdarian} were able to achieve a 74\% localization accuracy and a 94\% detection accuracy. \cite{kojuri} developed two different kinds of artificial neural networks-a radial basis function (RBF) and a multi-layer perceptron (MLP)-for the detection of myocardial infarction. \cite{sun} argue that supervised learning techniques have achieved limited success and apply multiple instance learning (MIL) to automatic electrocardiogram classification. They demonstrate that they're proposed algorithm called LTMIL surpasses previous supervised learning approaches in terms of classification quality. 

\cite{liu1} propose a novel electrocardiogram feature which they obtain by best 20th-order polynomial approximation of a given ECG signal. Developing a classification model using this feature, \cite{liu1} are able to achieve a 94.4\% accuracy for myocardial infarction detection. \cite{sharma} use stationary wavelet transform to decompose a given electrocardiogram signal into different sub-bands. Then, support vector machine (SVM) and k-nearest neighbor models are developed to classify the presence of myocardial infarction using sample entropy, normalized sub-band energy, log energy entropy, and median slope from the selected bands as features. \cite{kachuee} use transfer learning from a model trained for arrhythmia classification to develop a model capable of detecting myocardial infarction. A deep convolutional neural network model is used, achieving a 95.9\% classification accuracy for myocardial infarction detection. 

\cite{remya} use a simple adaptive threshold (SAT) to classify myocardial infarction. A multiresolution approach along with an adaptive thresholding is used to extract the depth of the Q peak and elevation in the ST segment which are the features used for classification. \cite{reasat} develop a subject-oriented approach based on a convolutional neural network which takes in leads II, III, and AVF. \cite{zewdie} model electrocardiogram signals using second order ordinary differential equations (ODE) and feed the best-fitting time varying coefficients of the ECG signal into a support vector machine (SVM) model. \cite{feng} develop a multi-channel convolutional-recurrent neural network model consisting of 16 convolutional layers followed by long-short term memory (LSTM) units to classify the presence myocardial infarction in electrocardiograms. Finally, \cite{liu2} propose a deep convolutional neural network which takes only 3-seconds of lead II as input at a time for myocardial infarction detection in ECG signals. 
 
Many approaches~\cite{safdarian, kojuri, sun, liu1, sharma, kachuee, remya, reasat, zewdie, feng} extract hand-crafted features from the raw ECG signals before feeding these features into a signal-processing algorithm. Such manual feature extraction is both time-consuming and not scalable. Our approach differs from others in the community in that it does not require any kind of manual feature extraction. Instead, we adopt a data-driven discovery approach in which we fully exploit the proven capabilities of deep learning algorithms to identify novel features or patterns in data that escape human notice or domain expertise, and which enable the use of neural network models to process data at scale. 

On the other hand, the few studies that do not do any feature engineering as a pre-processing step~\cite{acharya, liu2} do not provide any rationale for their choice of ECG leads to identify or classify heart conditions. To shed new information on this important decision making process, we quantify the contribution of each ECG and Frank lead to correctly identify heart conditions in ECG data. Our results indicate which lead(s) contain the most meaningful information for the detection of myocardial infarction. Based on these findings we design a neural network model that employs these leads to achieve state-of-the-art results.

\section{Methods}
  
We begin this section by describing the data curation methodology followed to prepare the training, validation and testing data sets. The ECG time-series data for this analysis was obtained from the public Physikalisch-Technische Bundesanstalt (PTB) database. Thereafter, we describe the architecture of the neural network model used to identify heart conditions. We provide all the key information for reproducibility purposes. All the code developed for this project is open source~\cite{heartrepo}.

\subsection{Data Curation}

As mentioned before, we use the PTB database, which consists of 549 ECG records from 290 unique patients, with a mean length of over 100 seconds for each record~\cite{ptbdb}. The dataset provides data from the 12 conventional ECG leads, along with 3 Frank leads, all sampled at 1000 Hz. Table~\ref{tab:zero} shows the distribution of heart conditions for the 290 patients in the dataset (Note: the diagnostic class was not available for 22 subjects; these subjects were not used in training.). This data set was used to train, validate and test our neural network models. 

\renewcommand{\arraystretch}{1.2}
\begin{table}[hbt!]
\centering
\caption{The diagnostic classes for unique patients in the dataset}
\begin{tabular}{ >{\centering\arraybackslash}m{1.9in} | >{\centering\arraybackslash}m{1.3in} }
\textbf{Diagnostic class}          & \textbf{Number of subjects} \\ \hline
Myocardial infarction & 148 \\ 
Cardiomyopathy/Heart failure & 18 \\
Bundle branch block & 15 \\
Dysrhythmia & 14 \\
Myocardial hypertrophy & 7 \\
Valvular heart disease & 6 \\
Myocarditis & 4 \\
Miscellaneous & 4 \\
Healthy controls & 52
\label{tab:zero}
\end{tabular}
\end{table}

\begin{figure*}
    \centering
    \begin{subfigure}[b]{0.45\textwidth}
        \centering
        \includegraphics[width=\textwidth]{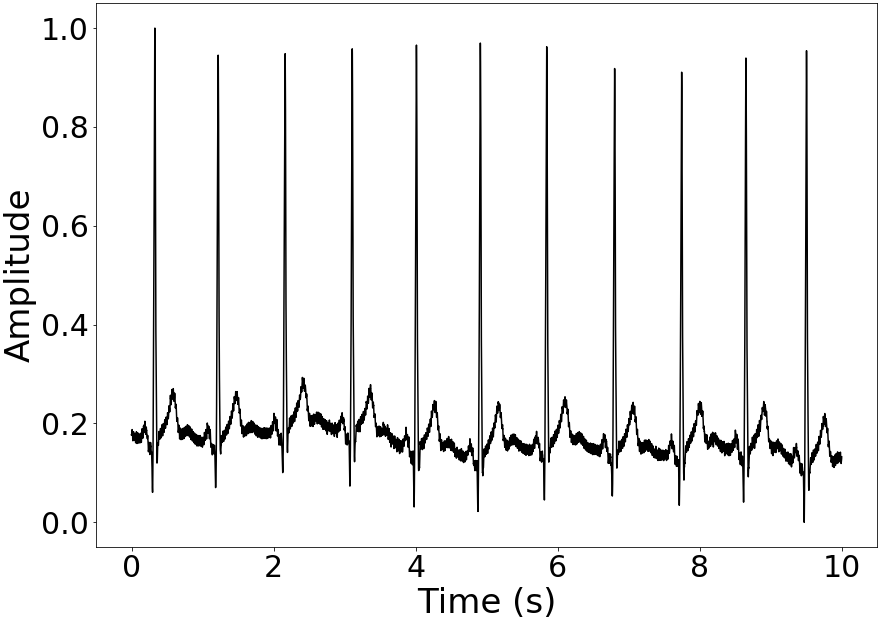}
        \caption[Network2]%
        {Channel v6 of healthy patient}    
        \label{fig:mean and std of net14}
    \end{subfigure}
    \hfill
    \begin{subfigure}[b]{0.45\textwidth}  
        \centering 
        \includegraphics[width=\textwidth]{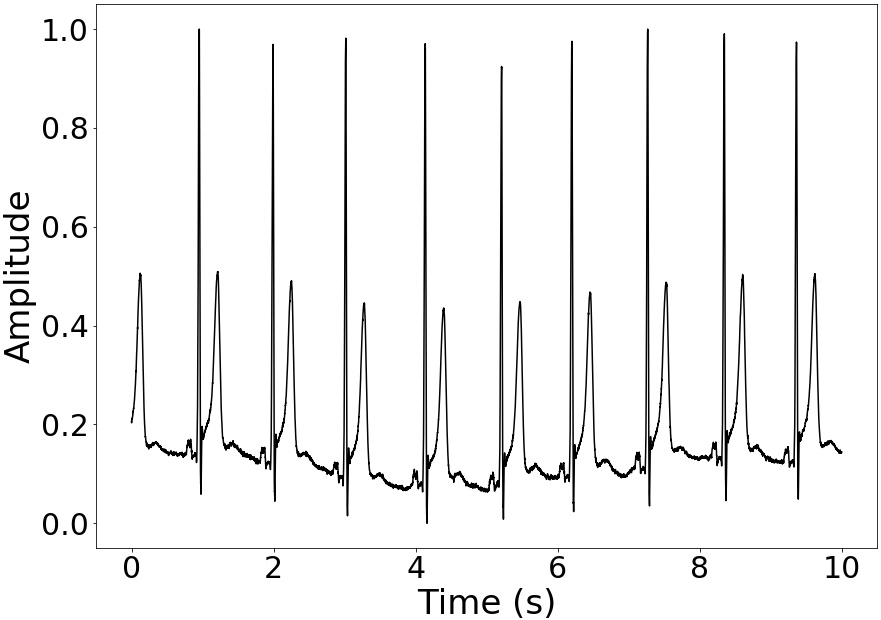}
        \caption[]%
        {Channel v6 of unhealthy patient}    
        \label{fig:mean and std of net24}
    \end{subfigure}
    \vskip\baselineskip
    \begin{subfigure}[b]{0.45\textwidth}   
        \centering 
        \includegraphics[width=\textwidth]{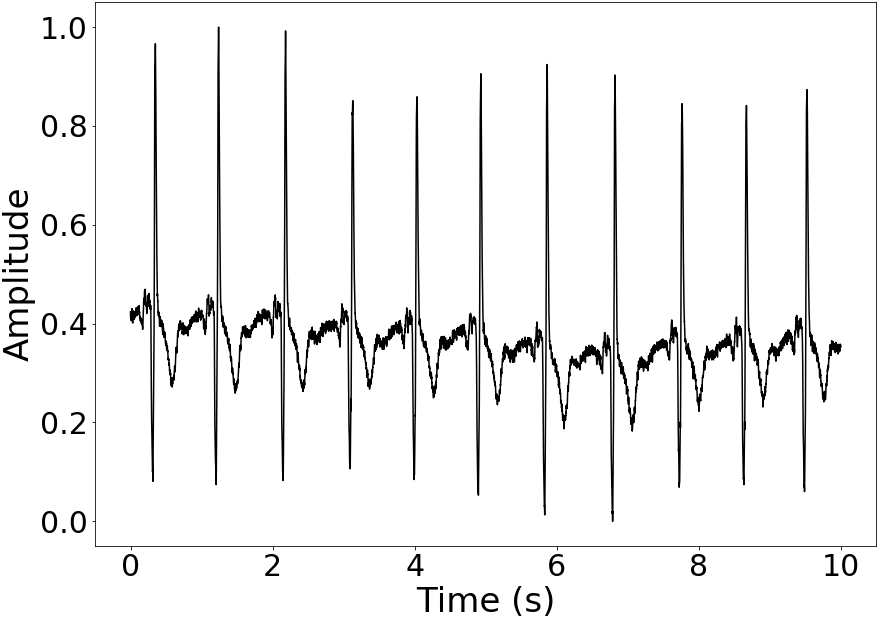}
        \caption[]%
        {Channel vz of healthy patient}    
        \label{fig:mean and std of net34}
    \end{subfigure}
    \hfill
    \begin{subfigure}[b]{0.45\textwidth}   
        \centering 
        \includegraphics[width=\textwidth]{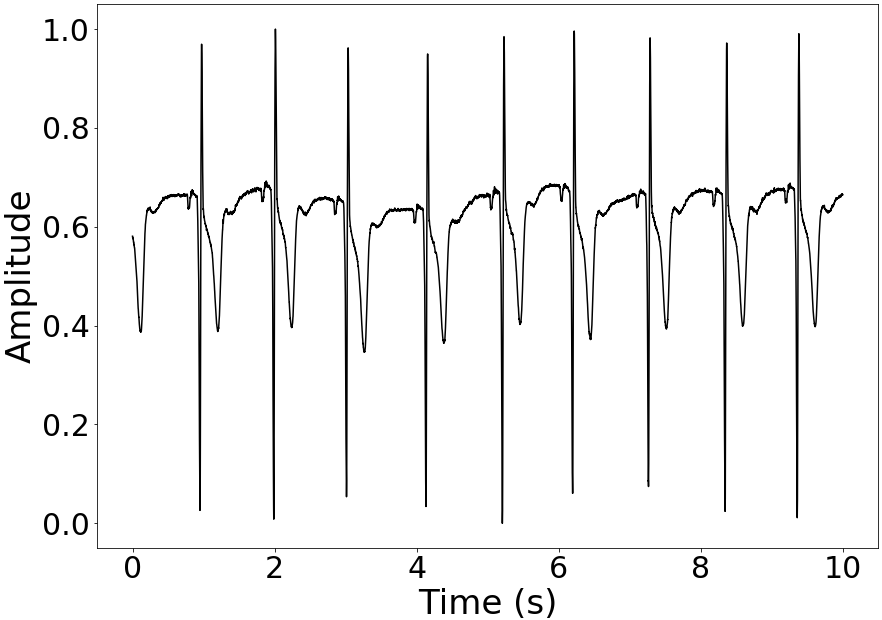}
        \caption[]%
        {Channel vz of unhealthy patient}    
        \label{fig:mean and std of net44}
    \end{subfigure}
    \caption[ The average and standard deviation of critical parameters ]
    {\small Sample input data for (a) and (c) a healthy patient, and (b) and (d) for an unhealthy patient} 
    \label{fig:mean_and_std_of_nets}
\end{figure*}

During the training stage, a 10-second long two-channel input was fed into the neural network. Both channels were normalized to ensure that the two channels were weighted equally, and time invariance was incorporated by selecting the 10 second long segment randomly from the entire signal. Figure~\ref{fig:mean_and_std_of_nets} shows sample inputs of the v6 and vz leads for patients with and without heart conditions.


\subsection{Neural network model}

An aspiration in deep learning research is to develop commodity software that may be utilized across disciplines that share common computational grand challenges. In this respect, deep transfer learning has had a number of successful applications~\cite{KHAN2019248,Tang:2019MNRAS,dglitcha:2017,weistarclusters,Ackerman2018MNRAS}.

\begin{figure}[!h]
    \centering
    \includegraphics[width=11cm]{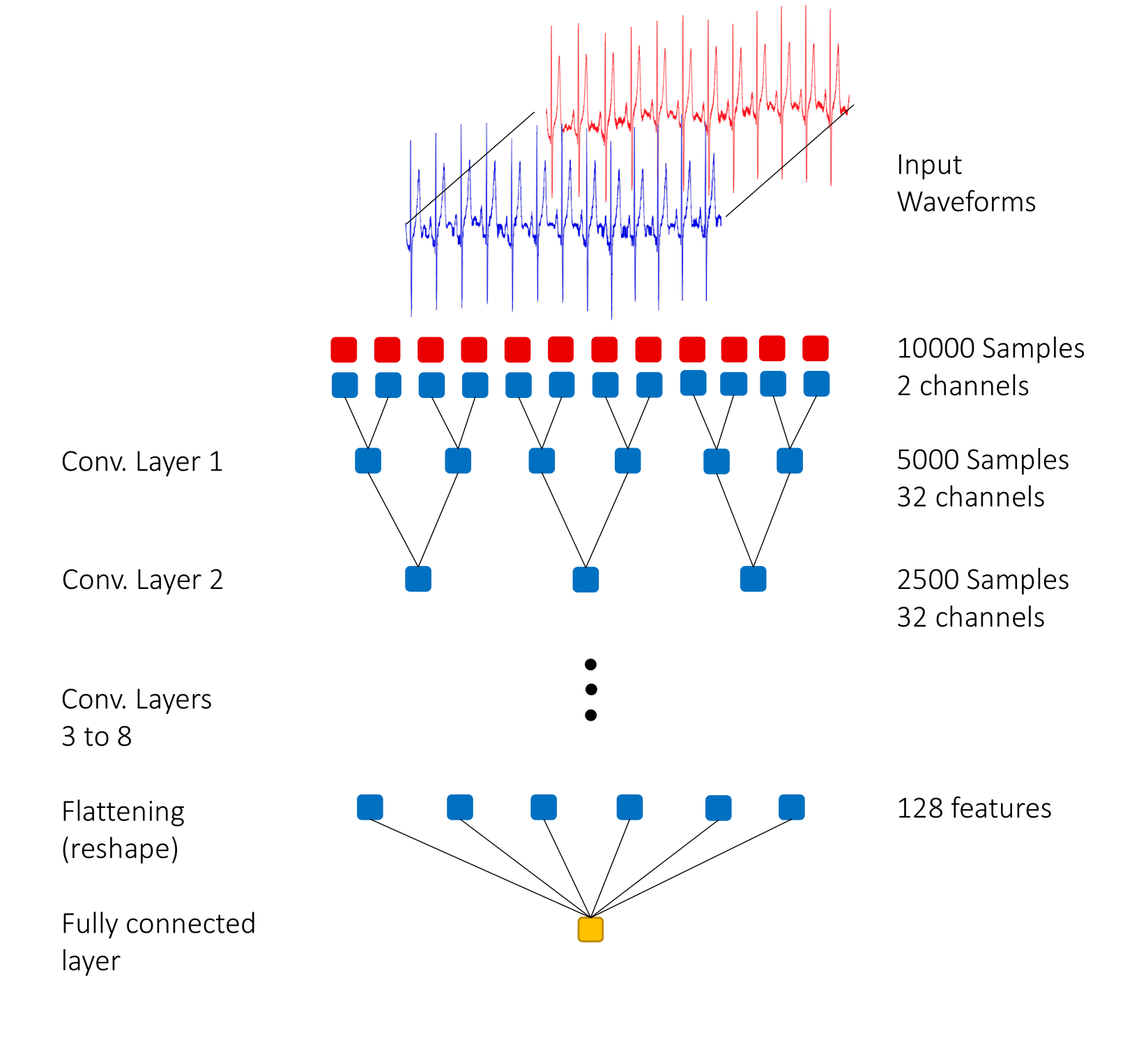}
    \caption{Architecture of the neural network model used to identify heart conditions feeding 10 seconds of raw ECG data from the v6 and vz ECG leads. Notice that these two independent time-series data sets are marked by the red and blue input data sets at the top of the diagram. The result is an activation of one of two classes, shown in yellow.}
    \label{fig:architecture}
\end{figure}

Our analysis explores the adequacy of this paradigm by adapting a neural network model that was originally designed to identify earthquakes by processing multiples channels of input data~\cite{cNQ}. Herein, we fine-tune the ConvNetQuake, using an eight-layer convolutional neural network architecture, shown in Figure~\ref{fig:architecture}, which processes 10-second long ECG signals to detect myocardial infarction. 

The primary difference between our architecture and that of ConvNetQuake is the use of a one-dimensional batch normalization layer after each convolutional layer to combat over-fitting. While there is much debate about whether batch normalization should proceed or follow the activation function, we observe that for our case, applying batch normalization after the activation yields better results. Another difference from ConvNetQuake is that we employ label smoothing. Label smoothing refers to the act of relaxing the confidence on the labels and is known to help discourage the model from making over-confident predictions. Our experiments showed that both of these techniques helped increase the accuracy of our models.

For training, we used a batch size of $10$ and a learning rate of $10^{-4}$. The weights were randomly initialized in all cases, and the ADAM optimizer was used. The two classes--myocardial infarction and healthy--are unbalanced in the database. To overcome this, we sampled input data such that the neural network was exposed to an equal amount of samples from each class. An 80-10-10 train-validation-test split was used in all instances. The analysis was carried out using NVIDIA V100 GPUs, with each experiment taking a few hours to train when trained on one GPU. 

It is known that random initialization of the weights at each training round may produce some variance in the performance of neural network models. In view of this observation, we have trained our models from scratch over 100 times to provide a fair representation of the performance of our models. In the results section, we provide average, mean and variance results of these training sweeps. Furthermore, to facilitate reproducibility of our results, we provide all the code used for these studies~\cite{heartrepo}.

\section{Results} 
\label{sec:results}

Here, we evaluate the performance of our models in myocardial infarction classification using standard statistical measures that others have used in the past: the accuracy, sensitivity, specificity, and recall of the models.

\subsection{Record-wise Split}

The first stage of our analysis consists of quantifying the contribution of each of the 15 leads---first individually, and then in pairs---to see which lead contains the most meaningful information for the detection of myocardial infarction. 

Tables~\ref{tab:one} and~\ref{tab:two} show our results obtained over a 20-fold cross-validation.

\renewcommand{\arraystretch}{1.2}
\begin{table}[hbt!]
\centering
\caption{Quantification of accuracies for \\single channels [i - v2] on PTBDB}
\begin{tabular}{c|c|c|c|c|c|c|c|c}
\textbf{Lead}          & i  & ii & iii & avl & avr & avf  & v1 & v2\\ \hline
\textbf{Average (\%)} & 82.68 & 89.32 & 84.91  & 84.84  & 86.96 & 87.44 & 87.15 & 83.93 \\ 
\textbf{Std. Dev. (\%)} & 6.74 & 5.52 & 6.48  & 3.12  & 5.23 & 4.75 & 4.55 & 6.38 \\ 
\textbf{Median (\%)} & 83.99 & 90.57 & 86.56  & 85.11  & 88.40 & 86.94 & 86.95 & 82.70
\label{tab:one}
\end{tabular}
\end{table}

\renewcommand{\arraystretch}{1.2}
\begin{table}[hbt!]
\centering
\caption{As Table \ref{tab:one}, for channels [v3 - vz]}
\begin{tabular}{c|c|c|c|c|c|c|c}
\textbf{Lead}         & v3 & v4 & v5  & v6 & vx & vy & vz\\ \hline
\textbf{Average (\%)} & 83.40  & 84.27 & 91.13 & 90.84 & 89.64  & 87.08  & 89.41 \\ 
\textbf{Std. Dev. (\%)} & 4.58  & 5.37 & 6.30 & 4.68 & 4.83  & 5.54  & 4.49 \\ 
\textbf{Median (\%)} & 82.38  & 84.55 & 91.84 & 92.04 & 89.28  & 88.38  & 90.33
\label{tab:two}
\end{tabular}
\end{table}

\noindent Based on these results, we observed that the five channels with the most valuable information for myocardial infarction detection were: v5, v6, vx, vz, and ii.

Next, we formed pairs of channels using the five best channels according to the PTB database and retrained the network with these pairs as input. Our results, presented in Tables~\ref{tab:four} and~\ref{tab:five} are based on a 20-fold cross validation.

\renewcommand{\arraystretch}{1.2}
\begin{table}[hbt!]
\centering
\caption{Quantification of accuracies for \\pairs of channels}
\begin{tabular}{c|c|c|c|c|c}
\textbf{Leads}          & v5, v6  & v5, vx & v5, vz & v5, ii & v6, vx \\ \hline
\textbf{Accuracy (\%)} & 92.26 & 93.77 & 93.46 & 94.19  & 92.98  \\ 
\textbf{Std. Dev. (\%)} & 4.78 & 4.30 & 4.20 & 4.85 & 3.63  \\ 
\textbf{Median (\%)} & 92.19 & 93.51 & 93.545 & 93.545 & 92.74
\label{tab:four}
\end{tabular}
\end{table}

\renewcommand{\arraystretch}{1.2}
\begin{table}[hbt!]
\centering
\caption{As Table \ref{tab:four}, continued}
\begin{tabular}{c|c|c|c|c|c}
\textbf{Leads}          & v6, vz & v6, ii & vx, vz & vx, ii & vz, ii \\ \hline
\textbf{Accuracy (\%)} & 94.76 & 93.58 & 93.44 & 93.35 & 92.17  \\ 
\textbf{Std. Dev. (\%)} & 3.87 & 5.09 & 3.99 & 4.63 & 4.23  \\ 
\textbf{Median (\%)} & 94.57 & 93.01 & 93.42 & 93.58 & 91.62 
\label{tab:five}
\end{tabular}
\end{table}

\noindent Tables~\ref{tab:four} and~\ref{tab:five} indicate that when lead v6 and lead vz are paired and fed into the neural network as a 2-channel input, the model is most successful at the task at hand. To see get a more precise estimate of the statistics, we trained a model on the best performing pair of channels 100 times (100-fold cross-validation). Table~\ref{tab:six} summarizes our findings.

\renewcommand{\arraystretch}{1.2}
\begin{table}[hbt!]
\centering
\caption{Results based on a 100-fold cross-validation}
\begin{tabular}{ >{\centering\arraybackslash}m{1.2in} | >{\centering\arraybackslash}m{2in} }
\textbf{Statistics}          & v6, vz \\ \hline
\textbf{Average (\%)} & 94.67 \\ 
\textbf{Std. Dev. (\%)} & 4.04 \\ 
\textbf{Median (\%)} & 94.67 
\label{tab:six}
\end{tabular}
\end{table}

\noindent As evident, the standard deviation across the 100-fold cross-validation is non-trivial. To figure out whether the standard deviation result was caused by a few outliers, which we confirmed to be the case, we calculated the statistics of the top 20 and top 50 performing models. We present these results in Table~\ref{tab:seven}. 

\renewcommand{\arraystretch}{1.2}
\begin{table}[hbt!]
\centering
\caption{Results based on best performing models}
\begin{tabular}{ >{\centering\arraybackslash}m{1.2in} | >{\centering\arraybackslash}m{1in} | >{\centering\arraybackslash}m{1in} }
\textbf{Statistics}          & Top-50 & Top-20 \\ \hline
\textbf{Average (\%)} & 97.91 & 99.43 \\ 
\textbf{Std. Dev. (\%)} & 1.60 & 0.47 \\ 
\textbf{Median (\%)} & 98.04 & 99.50 
\label{tab:seven}
\end{tabular}
\end{table}

\noindent As evident from Table~\ref{tab:seven}, the standard deviations dropped considerably relative to the standard deviation across all 100 models (see Table~\ref{tab:six}). The much lower standard deviation provides evidence for consistent strong performance; many models performed very well, but the overall statistics across 100 models were dragged down by a few models which performed very poorly.

In all of our cross-validation experiments, there are two factors varying across each iteration: the random initialization of the weights of the neural network, and the random train-val-test split. To investigate whether some models were performing poorly due to a poor train-val-test split\footnote{A poor train-val-test split refers to one where the distribution of the train data does not match the distribution of the test data. With such a small dataset, this occurs at a relatively high frequency when the data is split up randomly.} as opposed to an unlucky random initialization of weights, we trained the model 100 times on one specific train-val-test split that had done well once in a previous iteration. Table~\ref{tab:eight} contains these results.

\renewcommand{\arraystretch}{1.2}
\begin{table}[hbt!]
\centering
\caption{Results based on 100 iterations with a fixed train-val-test split}
\begin{tabular}{ >{\centering\arraybackslash}m{1.2in} | >{\centering\arraybackslash}m{2in} }
\textbf{Lead}          & v6, vz \\ \hline
\textbf{Average (\%)} & 99.89 \\ 
\textbf{Std. Dev. (\%)} & 0.43 \\ 
\textbf{Median (\%)} & 100
\label{tab:eight}
\end{tabular}
\end{table}

\noindent Through this exercise, we found a few train-val-test split cases in which the classification accuracy was suboptimal. The results on train-val-test splits with different distributions are poor because of the nature of the splits and are unable to assess the model’s capabilities, which is what we filter out. This could be remedied with access to a larger dataset to further improve the resilience and robustness of our neural network model; we hope to investigate this in the near future. A key goal of this project is to engage with the cardiology community to further improve the performance of the models introduced herein by getting access to larger data banks, and to design ready-to-use apps based on deep learning algorithms for their use in realistic diagnostics settings.

To put our results in context, we have compiled contemporaneous results in the literature, which have utilized the same dataset, along with ours in Table~\ref{tab:nine}. 

\renewcommand{\arraystretch}{1.3}
\begin{table}[hbt!]
\centering
\caption{Comparison of our results with other studies in the literature that also use the PTB database. Best results are marked in boldface.}
\begin{tabular}{c|c|c|c|c}
 \textbf{Work} & \textbf{Accuracy (\%)} & \textbf{Sensitivity (\%)} & \textbf{Specificity (\%)} & \textbf{Precision (\%)} \\ \hline
 Acharya et al. & 93.5 & 93.7 & - & 92.8 \\ 
 Safdarian et al. & 94.7 & - & - & - \\ 
 Kojuri et al. & 95.6 & 93.3 & - & 97.9 \\ 
 Sun et al. & - & 92.6 & - & 82.4 \\ 
 Liu et al. & 94.4 & - & - & - \\ 
 Sharma et al. & 96 & 93 & - & 99 \\ 
 Kachuee et al. & 95.9 & 95.1 & - & 95.2 \\ 
 Remya et al. & 93.61 & 93.22 & 94.28 & - \\
 Reasat et al. & 84.54 & 85.33 & 84.09 & - \\
 Zewdie et al. & 98.3 & 98.7 & 96.4 & - \\
 Feng et al. & 95.4 & 98.2 & 86.5 & - \\
 Strodthoff et al. & - & 93.3 & 89.7 & - \\
 Huang et al. & 96.96 & \textbf{99.89} & 92.51 & 95.35 \\
 Liu et al. & 98.59 & 99.53 & 94.50 & - \\
 Ours & \textbf{99.43} & 99.40 & \textbf{99.45} & \textbf{99.46}
 \label{tab:nine}
 \end{tabular}
\end{table}

\subsection{Patient-wise Split}

The dataset used contained 549 records from 290 unique patients. This indicates that some patients had multiple records (i.e., multiple diagnoses). It is important to note that these diagnoses could have had different outcomes (a given patient could be healthy in one record and unhealthy in another). The analysis conducted above was done on a record-wise split, meaning that if a record of a particular patient was in the train set, a different record of that same patient could show up in the validation or test set. We investigated how our results would change if we instead did a patient-wise split; that is, all records of a particular patient show up only in either the train, validation, or test split. The procedure used for a record-wise split was replicated for a patient-wise split -- the results for the best performing pairs of channels are presented in tables \ref{tab:ten} and \ref{tab:eleven}.

\renewcommand{\arraystretch}{1.2}
\begin{table}[hbt!]
\centering
\caption{Quantification of accuracies for \\pairs of channels (patient-wise split)}
\begin{tabular}{c|c|c|c|c|c|c}
\textbf{Leads}          & i, vz  & ii, v1 & ii, vz & avr, v6 & avf, vz & v1, v6\\ \hline
\textbf{Accuracy (\%)} & 93.93 & 92.94 & 95.73 & 94.51 & 93.01 & 91.76 \\ 
\textbf{Std. Dev. (\%)} & 0.96 & 1.33 & 1.19 & 0.84 & 1.17 & 1.59 \\ 
\textbf{Median (\%)} & 93.91 & 92.89 & 95.59 & 94.54 & 92.77 & 92.08
\label{tab:ten}
\end{tabular}
\end{table}

\renewcommand{\arraystretch}{1.2}
\begin{table}[hbt!]
\centering
\caption{As Table \ref{tab:ten}, continued}
\begin{tabular}{c|c|c|c|c|c|c}
\textbf{Leads}           & v4, vz & v5, v6 & v5, vx & v6, vx & v6, vz & vy, vz\\ \hline
\textbf{Accuracy (\%)}  & 93.18 & 94.15 & 94.45 & 94.28 & \textbf{96.47} & 94.33 \\ 
\textbf{Std. Dev. (\%)}  & 0.90 & 0.90 & 0.73 & 1.12 & \textbf{0.51} & 0.91  \\ 
\textbf{Median (\%)}  & 92.91 & 94.69 & 94.70 & 94.26 & \textbf{96.23} & 94.17 
\label{tab:eleven}
\end{tabular}
\end{table}

As can be seen from the tables \ref{tab:ten} and \ref{tab:eleven}, the best performing pair of channels with a patient-wise split is (v6, vz), which happens to be the same as what we discovered for a record-wise split. The second best pair, (ii, vz) also contains the vz channel. We decided to see if combining these three channels (v6, vz, ii) would give us better results. After conducting a study of these three channels with various train-val-test splits, we found that this combination does yield better results. The following table presents our results over 10 trials:

\renewcommand{\arraystretch}{1.2}
\begin{table}[hbt!]
\centering
\caption{Results for three channels: (v6, vz, ii)}
\begin{tabular}{ >{\centering\arraybackslash}m{1.2in} | >{\centering\arraybackslash}m{2in} }
\textbf{Lead}          & v6, vz, ii \\ \hline
\textbf{Average (\%)} & 97.83 \\ 
\textbf{Std. Dev. (\%)} & 0.56 \\ 
\textbf{Median (\%)} & 97.92
\label{tab:twelve}
\end{tabular}
\end{table}

As evident, using a combination of three channels is better than using pairs of channels. The patient-wise split results obtained here are lower than the results obtained for a record-wise split -- this is expected as in a record-wise split, there may be some leakage from the train data into the test data. The other studies we compared our results to in Table~\ref{tab:nine} do not explicitly mention whether a record-wise or patient-wise split was used. 

\section{Discussion} 

This paper introduces a new architecture for heart condition classification based on raw ECG signals using multiple leads. Our approach outperforms the current state-of-the-art model on this dataset by a large margin of almost one full percent, with no manual feature engineering.

Another way in which our approach differs from others in the community is in our choice of leads. While other works in this domain have employed simply one or two of the 15 leads that the PTB database provides for each record, strong justification for their choice of lead(s) hasn't been provided.

Here, we studied which of the 15 ECG channels (12 conventional ECG leads and 3 Frank leads) contains the most meaningful information with respect to myocardial infarction detection, finding that channels v6, vz and ii are the most significant. Our work also illustrates that recent advances in machine learning can be leveraged to produce a model capable of classifying myocardial infarction with a cardiologist-level success rate.

\paragraph{Limitations}

As in many other deep learning studies in the literature, one of the key limitations to maximize the power of neural network models is the lack of labelled data~\cite{weistarclusters}. We also face a similar challenge in this study, and candidly acknowledge that to further improve the performance of our models we require access to a larger, labelled data set. With a small database such as the PTB database, it is difficult to gauge the true robustness of the model since the test set ends up being relatively small. We are in the search for a larger data set to further increase the accuracy and generalization of our models, e.g., to correctly identify other heart conditions beyond myocardial infarction. If that were accomplished, then we would readily share our deep learning models and develop an open source, computationally efficient app that may be readily used by cardiologists.

\section{Conclusion} 

We have conducted a detailed analysis of the relative importance of each of the standard 15 ECG channels to identify myocardial infarction using a multi-channel neural network model. Our findings indicate that deep learning can identify this heart condition upon processing only ten seconds of raw ECG data from the v6, vz and ii leads, reaching cardiologist-level success rate.

Through this analysis we: (i) furnished evidence that deep learning algorithms may be readily used as commodity software, i.e., taking two disparate areas of research---cardiology and seismology---we showed that a neural network model that was originally designed to identify earthquakes by processing multiple input channels may be re-designed and tuned to identify myocardial infarction; (ii) deep learning does no require feature engineering of ECG data to identify myocardial infarction in the PTB database. Indeed, our deep learning model only requires ten seconds of raw ECG data to identify this heart condition with cardiologist-level performance.

We also recommend to provide deep learning researchers access to a larger database to further improve and extend  to other types of heart conditions the results presented in this manuscript.  We look forward to working with the cardiology community to develop deep learning algorithms that may be readily applicable in a diagnostics setting, providing trustworthy, real-time information regarding heart conditions with minimal computational resources that are either deployed on the cloud or in lightweight, personalized devices. 

\vspace{3mm}

\textbf{Acknowledgments.} EAH and ZZ gratefully acknowledge National Science Foundation (NSF) awards OAC-1931561 and OAC-1934757. AG acknowledges support from the Fiddler Innovation Undergraduate Fellowship and the Students Pushing Innovation (SPIN) program at the National Center for Supercomputing Applications at the University of Illinois at Urbana-Champaign.

This work utilized resources supported by the NSF's Major Research Instrumentation program, grant OAC-1725729, as well as the University of Illinois at Urbana-Champaign. We are grateful to NVIDIA for donating several Tesla P100 and V100 GPUs that we used for our analysis, and the NSF grants NSF-1550514, NSF-1659702 and TG-PHY160053. This research used resources of the Argonne Leadership Computing Facility, which is a DOE Office of Science User Facility supported under Contract DE-AC02-06CH11357. We thank the NCSA Gravity Group for useful feedback. 

%
%








\printbibliography

\end{document}